\documentclass{article}

\usepackage[preprint,nonatbib]{neurips_2019}

\usepackage[utf8]{inputenc} 
\usepackage[T1]{fontenc}    
\usepackage{hyperref}       
\usepackage{url}            
\usepackage{booktabs}       
\usepackage{amsfonts}       
\usepackage{nicefrac}       
\usepackage{microtype}      

\usepackage[pdftex]{graphicx}
\usepackage{algorithm,algcompatible,amsmath}
\algnewcommand\INPUT{\item[\textbf{Input:}]}%
\algnewcommand\OUTPUT{\item[\textbf{Output:}]}%
\usepackage{color}

\title{A Bulirsch-Stoer algorithm using Gaussian processes}

\author{%
  Philip G.~Breen \\
  School of Mathematics\\
  Univeristy of Edinburgh\\
  Edinburgh, EH9 3FD, UK \\
  \texttt{phil.breen@ed.ac.uk} \\
   \And
Christopher N.~Foley \\
  MRC Biostatistics Unit\\
  University of Cambridge\\
  Cambridge, CB2 0SR, UK \\
  \texttt{christopher.foley@mrc-bsu.cam.ac.uk} 
}

\begin{document}

\maketitle

\begin{abstract}
In this paper, we treat the problem of evaluating the asymptotic error in a numerical integration scheme as one with inherent uncertainty. Adding to the growing field of probabilistic numerics, we show that Gaussian process regression (GPR) can be embedded into a numerical integration scheme to allow for (i) robust selection of the adaptive step-size parameter
and; (ii) uncertainty quantification in predictions of putatively converged numerical solutions. We present two examples of our approach using Richardson's extrapolation technique and the Bulirsch-Stoer algorithm. In scenarios where the error-surface is smooth and bounded, our proposed approach can match the results of the traditional polynomial (parametric) extrapolation methods. In scenarios where the error surface is not well approximated by a finite-order polynomial, e.g.  in the vicinity of a pole or in the assessment of a chaotic system, traditional methods can fail, however, the non-parametric GPR approach demonstrates the potential to continue to furnish reasonable solutions in these situations. 
\end{abstract}

\section{Introduction}

Solutions to ordinary differential equations (ODEs) are often sought by means of adaptive numerical integration algorithms \cite{NumRce}. The accuracy of these approaches regularly relies on the assumption that the {\sl true} solution is suitably smooth across the region of integration so that {\sl locally}, i.e. in an interval of width equal to integration step-size $h_{i}$, where $i$ indexes the algorithm's iteration, the numerical error is well approximated by a finite-order polynomial. The choice of  step-size $h_{i}$ is then determined using an extrapolation procedure and a given precision bound. Of particular importance to the present study are the extrapolation procedures of Richardson \cite{Rich} and the Bulirsch-Stoer algorithm  (BSA) \cite{BS1964}. The BSA is a descendant of Richardson's algorithm and aims to quantify whether the numerical integration scheme has recovered a converged solution to the ODE at the $i^{th}$ iteration by recursively assessing the behaviour of the numerical error in the limit as $h_{i}\to 0$. Central to the good performance of this method is the assumption that the error function can be expressed as a power series in $h_{i}$. While this assumption has broad applicability, it can breakdown in a number of scenarios, e.g. in the vicinity of a singularity or when aiming to compute particle trajectories in chaotic systems - owing to exponentially large separation rates between nearby particles. In scenarios like these, numerical integration algorithms which are underpinned by polynomial extrapolation can fail and often with considerable burden on computational economy.
 
In this paper we instead propose treating the numerical error as {\sl uncertain}. We employ a probabilistic reinterpretation of the problem of adaptively determining a suitable choice of step-size $h_{i}$ at each iteration. We do this by embedding Gaussian process regression (GPR) in to the extrapolation component of the numerical integration scheme. Our non-parametric approach side-steps the assumption that the numerical error is `well-behaved' near the region of interest and contrasts with the parametric (polynomial) extrapolation methods of Richardson and Bulirsch-Stoer. Moreover, the proposed approach allows for uncertainty quantification in the choice of step-size, at each iteration, and consequently the numerical error. Hence, the GPR implementation can allow for a broader theoretical assessment of candidate numerical solutions to the ODE relative to alternative {\sl deterministic} methods. Our GPR extrapolation method contributes to the expanding field of probabilistic numerics \cite{BOGO2018,HOG2015,Cockayne2017} and as we will demonstrate may offer a more robust numerical integration scheme, particularly when the true solution to the ODE over the region of interest is not well approximated by a finite-order polynomial.

We begin our discussion by considering the widely used extrapolation scheme put forward by Richardson. After a brief mathematical description of the method, we introduce how a GP can be successfully included into the scheme and present an algorithm to do this. The Bulirsch-Stoer algorithm is a descendant of Richardson's scheme and is regularly used to compute practical numerical solutions to ODEs. We summarize the BSA before presenting a probabilistic version of the algorithm with a application. We finish by applying both the probabilistic and deterministic versions of the algorithm to the classical problem of a Keplerian orbit.

Our goal is to approximate the integral $F^{\ast} = \int_{\Omega} f(x) dx $, which we assume exists and is bounded over the domain $\Omega$. Where possible, we avoid unnecessary mathematical details and instead focus on describing the various issues associated with the computation of $F^{\ast}$, i.e. owing to the behaviour of the integrand $f(x)$, either locally or globally in $\Omega$, or because of the choice of numerical integration scheme.

\section{Richardson extrapolation}

Let $I\left(f, h\right)$ be a numerical evaluation of the integral $\int f(x) dx$ using a fixed step-size $h$. Generally, if $f(x)$ is smooth and bounded across $x\in \Omega$, $I\left(f, h\right)$ can be written as a weighted sum of $N$ evaluations of $f(x)$ over a sub-interval $\omega\subseteq \Omega$ of length $h$, i.e.
\[
I\left(f, h\right) = h \sum_{i=1}^{N}  w_i f(x_i) \approx \int_{\omega\subseteq\Omega} f(x)  dx,
\]
where the weights $w_{i}$ and number of abscissa $N$ are given by the choice of numerical scheme, e.g. a Gaussian or Netwon-Cotes type quadrature method. As it will be important later, the trapezoidal rule, from the Newton-Cotes family of numerical schemes, sets $N=2$, $w_{1}=w_{2} = 0.5$, $h = x_{2}-x_{1}$. In all evaluations of this type, the error in approximating the integral by $I\left(f, h\right)$ is particularly sensitive to the choice of step parameter $h$.

Under the assumption that the numerical evaluation can be written as
\[
I\left(f, h_{0}\right) = F^{\ast} + C_{0}h_{0}^{n} + \mathcal{O}\left(h_{0}^{n+1}\right), \quad n \in \mathbb{N}
\]
where $h_{0}$ is some choice of step parameter $h$ and $C_{0}$ a bounded constant, Richardson's extrapolation scheme can be used to: (i) assess sensitivity to the choice of $h$ and; (ii) recursively reduce the leading order approximation error, i.e. sequentially remove $C_{0}h_{0}^{n}$. Note, in order to simplify notation the above equation assumes that $\omega = \Omega$. To derive Richardson's extrapolation scheme, we re-write the numerical approximation in terms of a new step-size $h_{1}= \gamma h_{0}$, for some $0<\gamma<1$ (the choice of which is discussed later), and combine $I\left(f, h_{0}\right)$ and $I\left(f, h_{1}\right)$ to remove the leading order error term.  Repeating this process for $h_{j}=\gamma h_{j-1}, j=1,2,\dots$, the scheme can be written as a recurrence relation in which, for each additional iteration, the approximation error decreases\footnote{The error cannot be reduced beyond numerical rounding error}, i.e.
\[
R_{j} = \frac{\gamma^{1-n-j} I(f,h_{j}) - R_{j-1}}{\gamma^{1-n-j}-1} = F^{*} + C_{j}h_{0}^{n+j} + \mathcal{O}\left(h_{0}^{n+j+1}\right)\,: \quad j = 1,2 \dots \quad \text{and} \quad R_{0} = I(f,h_{0}).
\]
Ignoring numerical rounding errors, the approach can be used to compute $F^{\ast}$ to within a desired tolerance $\tau$, which we refer to as a {\sl converged} numerical solution, e.g.  by identifying $h_{j}:\,\,\mid R_{j} - R_{j-1} \mid \leq \tau$. For example, the method underpins the Romberg extension of the trapezoidal scheme. 

A variation of Richardson's method, sometimes referred to as Richardson's deferred approach to the limit, computes an ensemble of values $R = \{R_{1}, R_{2}, \dots, R_{J}\}$ and fits a polynomial, as a function of $h_{0}$, to $R$. The intercept of the fitted polynomial is then an estimate of $I(f, 0)$, i.e. $F^{\ast}$ in the theoretical limit as $h_{0}\to 0$, which we denote by $F^{\ast}_{h_{0}\to 0}$. The process is repeated for several distinct initializations of $h_{0}$, e.g. $h^{(1)}_{0}$ and $h^{(2)}_{0}$, and a converged numerical solution is identified when $\mid F^{\ast}_{h^{(1)}_{0}\to 0} - F^{\ast}_{h^{(2)}_{0}\to 0} \mid \leq \tau$. This approach underpins the Bulirsch-Stoer  \cite{BS1964} extension of Richardson's extrapolation technique.

Central to the good performance of both methods is the {\sl parametric} assumption that the numerical error can be approximated by a power series in $h_{0}$ of order $n$, or more generally on $f(x)$ being {\sl analytic} and bounded in $\Omega$. The assumption is widely applicability, but can fail when two nearby approximations, e.g. at $h_{j}$ and $h_{j+1}$, diverge at a rate not captured by $h_{0}^{n}$, i.e. near to a singular point or owing to the systems intrinsic chaotic nature. To overcome this limitation, we propose a non-parametric approach that is inspired, in part, by recent developments in the Bayesian Quadrature literature \cite{D1988,OH1991}. We illustrate the approach considering the example of Richardson's deferred approach to the limit, i.e. the Bulirsch-Stoer algorithm.
\subsection{Uncertainty in the numerical evaluation}
So far we have treated the numerical evaluation as one in which the approximation error is fixed by a given choice of step-size $h$. We now make a conceptual shift from this position and instead assume that the approximation error has intrinsic uncertainty. For a very good discussion of the philosophical and scientific implications of this switch see \cite{HOG2015}. In the present paper, we let $\hat{I}\left(f, h_{j}\right)$ denote the probabilistic re-interpretation of $I\left(f, h_{j}\right)$, where
\[
\hat{I}\left(f, h_{j}\right) = F^{\ast} + \epsilon_{j}(h_{j}), \quad F^{\ast} \in \mathbb{R} \quad \text{and} \quad h_{j}\in \mathbb{R}^{+},
\]
i.e. the uncertainty appears through the presence of the random error term $\epsilon_{j}$. By construct, the approximation error must approach zero as the step-size approaches zero, which, in the probabilistic model, requires that
\[
p\left(\hat{I}\left(f, h_{j}\right) - F^{\ast} = 0 \mid F^{\ast}\right) \to 1,  \quad j\to \infty \quad \text{and} \quad 0\leq h_{j}<h_{j-1}.
\]
The identity is motivated by a key assumption we make: that a numerical evaluation $I(f, h_{j})$, computed using a deterministic numerical integration scheme, is a draw from the distribution of probabilistic evaluations $\hat{I}(f, h_{j})$. Hence, as $\lim_{h\to0}\left(I(f, h) - F^{\ast}\right) = 0$, the above identity holds. We furthermore consider that a collection of $J$ random variables $\hat{I}(f, h_{j})\,:\,\,j = 1,2,\dots, J$ is a Gaussian process (GP), where $h_{j}$ denotes a {\sl distinct} step-size that monotonically decreases as $j$ increases. In the spirit of the Bulirsch-Stoer algorithm, our objective is to estimate $F^{\ast}$ at $h=0$ from the fitted estimate of $\hat{I}\left(f, 0\right)$. In the probabilistic reinterpretation, an estimate of $F^{\ast}$ is deemed a converged numerical evaluation when the uncertainty $\sigma_{0}$ in $\hat{I}\left(f, 0\right)$ is below a user defined uncertainty tolerance $\hat{\tau}$ (henceforth we drop the hat). Therefore, in deterministic evaluations of $F^{\ast}$, the tolerance $\tau$ places a bound on the numerical precision of an estimate, whereas in probabilistic evaluations $\tau$ places a bound on the maximum uncertainty of an estimate. Our approach can broadly be described in four steps.

Step 1; Input a $K$ dimensional vector of step-sizes ${\bf h} = \{h_{1}, h_{2}, \dots, h_{K}\}$ and compute the initial data set $\mathcal{D}^{(0)}_{h}$ comprising $j = 1,2,\dots,K$ pairs of variables $\{h_{j}, I(f,h_j)\}\in \mathcal{D}^{(0)}_{h}$, where each $I(f,h_j)$ is assumed a draw from the distribution of probabilistic evaluations $\hat{I}(f, h_{j})$. $I(f,h_j)$ is computed via a numerical integration scheme, and in the present treatment we use the trapezoidal scheme. Step 2; Let the collection of $K$ random variables $\hat{I}(f,h_{j}), j = 1,2,\dots,K,$ be a GP, i.e. 
\[
\hat{I}(f, {\bf{h}}) \mid \mathcal{D}^{(0)}_{h} \sim \mathcal{GP} \left(m({\bf{h}}), k\left({\bf{h}}, {\bf{h}}'\right)\right),
\]
where $m(.)$ and $k(.,.)$ denote the GP mean and kernel functions, respectively. From this model, compute estimates of: (i) $F^{\ast}$ at $h=0$, denoted $F^{\ast}_{0}$, and; (ii) $\sigma_{0}$, i.e. the uncertainty of in the estimate $F^{\ast}_{0}$. Step 3; if the uncertainty in the estimate is above a user defined threshold, i.e. $\sigma_{0}>\tau$, we reject $F_{0}^{\ast}$ as a reasonable estimate of $F^{\ast}$ and instead estimate a new step-size $h_{K+1}$ which satisfies
\[
h_{K+1} \,: \quad \sigma(h_{K+1}) = \hat{\gamma} \sigma_{0}, \quad 0<\hat{\gamma}<1,
\]
i.e. locate $h_{K+1}$ such that the uncertainty $\sigma(h_{K+1})$ in the fitted value $\hat{I}(f, h_{K+1})$ is a user defined fraction $\hat{\gamma}$ of the uncertainty in the current estimate $F_{0}^{\ast}$ of $F^{\ast}$. In practice $h_{K+1}$ is identified using a grid search over the region $h_{K+1}\in [0,h_{K})$. Step 4; augment the sample data with a new observation using $h_{K+1}$, i.e.
\[
\mathcal{D}^{(1)}_{h} = \mathcal{D}^{(0)}_{h} \cup \{h_{K+1}, I(f,h_{K+1})\},
\]
where $I(f,h_{K+1})$ is computed via the deterministic numerical integration scheme. Using the augmented data, we then repeat steps 2-4 until we identify an estimate of $F^{\ast}$ with uncertainty $\sigma_{0}\leq\tau$. See Alg. \ref{algRE} for pseudo-code. 

In summary, the proposed approach is an {\sl adaptive learning} technique, developed from Richardson's extrapolation and the Bulirsch-Stoer algorithm, that sees the uncertainty in the current estimate $F_{K+j}^{\ast}$ of $F^{\ast}$ being used to guide the choice of subsequent step-sizes $h_{k+j+1}$. There are two notable benefits of this approach over and above the deterministic alternative: (i) there is no parametric assumption concerning the behaviour of the error function, i.e. that the leading order error term is well approximated by of a finite order polynomial, and; (ii) the choice of step-size between two observations $I(f. h_{j})$ and $I(f, h_{j+1})$ is `dynamic', i.e. in the deterministic set-up the distance between two abscissa $h_{j}$ and $h_{j+1}$ is fixed and controlled by the parameter $\gamma$ whereas in the probabilistic setting the distance can vary between any two adjacent abscissa owing to asymmetry in the distribution of the uncertainty parameter $\sigma_{0}$ and the choice of $\hat{\gamma}$ (henceforth we drop the hat). A fundamental challenge remains between both the deterministic and probabilistic approaches, however, as each subsequent step-size $h_{K+i}$, $i=1,2,\dots,J$, must be small enough to closely approximate $I(f,h)$ while large enough that computation is dominated by {\sl discretization} error only, e.g. see \cite{HV2016}. In the probabilistic setting, this issue is related to the choice of step-control parameter $\gamma$, the choice of `prior' mean and the kernel function of the GP. 

\begin{figure}
    \centering
    \includegraphics[width=0.49\linewidth]{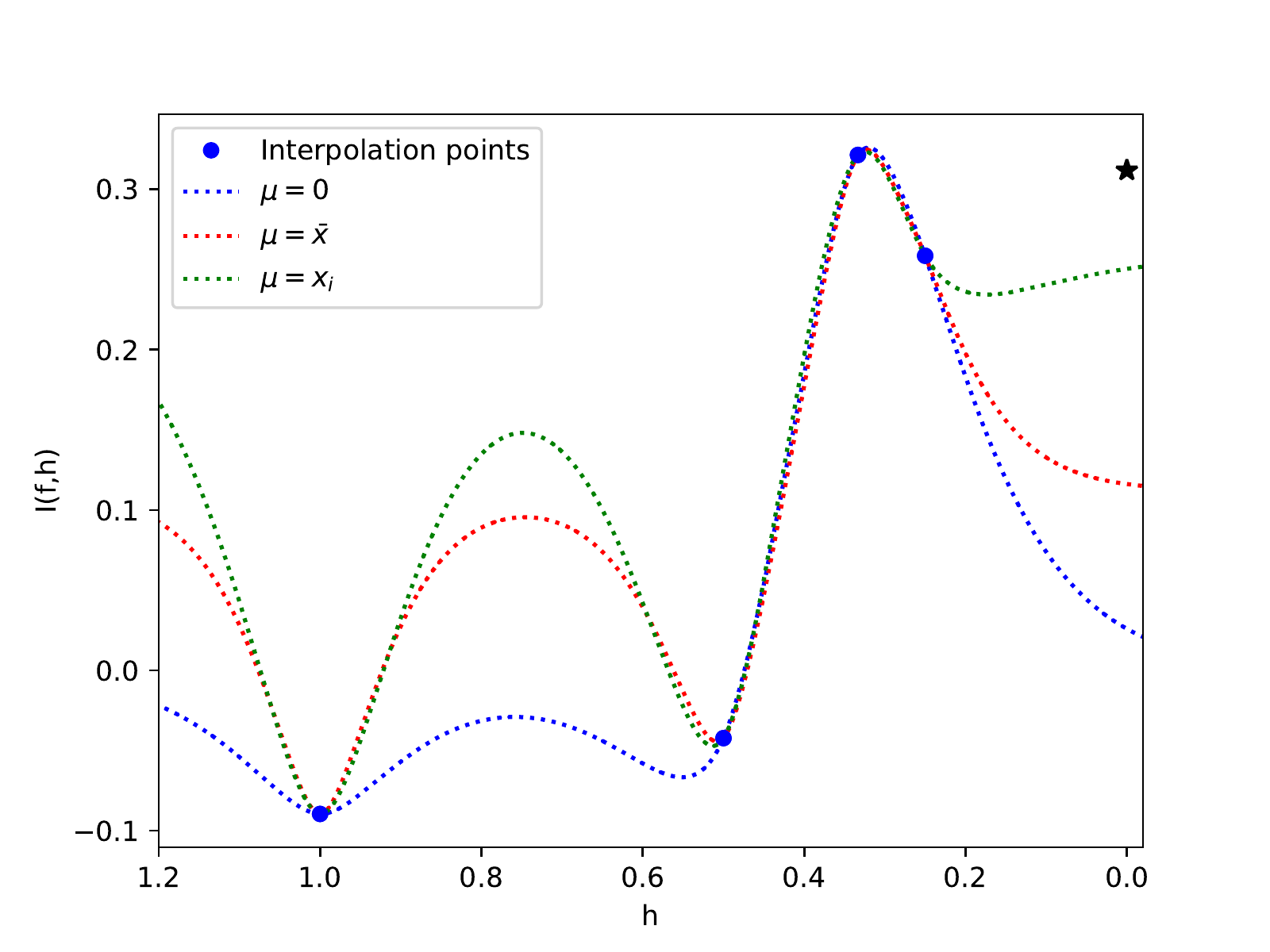}
    \includegraphics[width=0.49\linewidth]{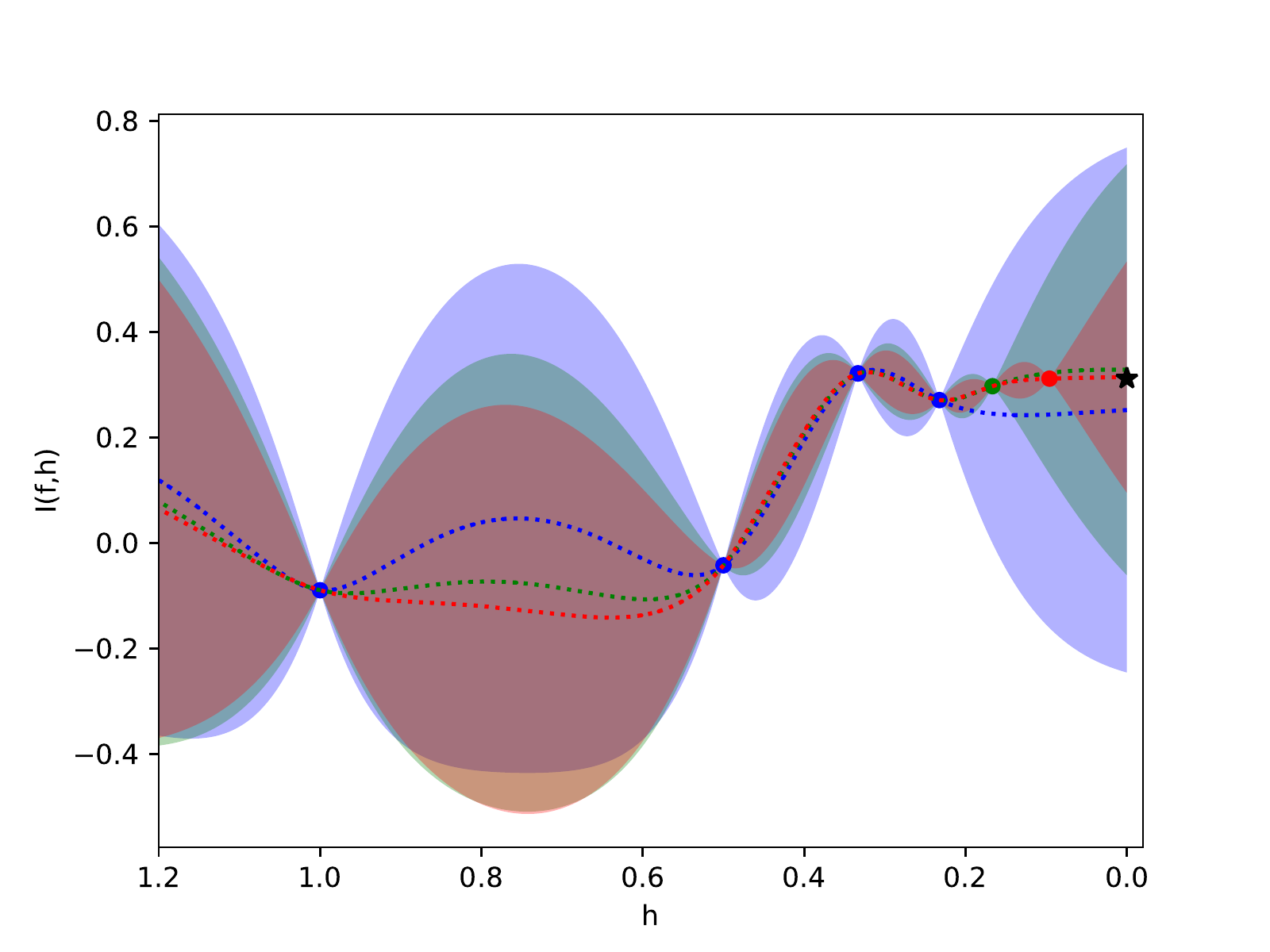}
    \caption{{\bf Left:} assessment of sensitivity to the choice of the prior mean ($\mu$) in  the $\mathcal{GP}$. Three choices are considered: $\mu=0$ , $\mu=\bar{x}$ and  $\mu=x_i$.
    {\bf Right:} Evaluation $F^{\ast}$ (marked with a star at $h=0$). The {\color{blue} Blue} points denote the evaluation of $I(f,h)$ using the 4 initial step-sizes ${\bf h}_{K} = \{1,1/2,1/3,1/4\}$ and a 95\% uncertainty band. We then apply the $\mathcal{GP}$ algorithm over several iterations (only 2 required) to identify a converged estimate of $F^{\ast}$. Each of the two new step-sizes were chosen based on the step-size $h_{j}$ in which the 95\% uncertainty band is below $\gamma = 0.5$ of the uncertainty $\sigma_{0}$ at $h=0$. The first evaluation, point $h_{5}$, is denoted by {\color{green} Green} point and the second, $h_{6}$, is denoted by {\color{red} Red}. 
   }
    \label{fig:exp}
\end{figure}

To illustrate our approach, and to go some way toward assessing the influence of these choices, we consider the following test function
\begin{equation}\label{eq:test_f}
f(x) = \exp \bigg( -\frac{(x - 0.35)^2}{2(0.1)^2} \bigg) - \frac{\sin(10x)}{3}
\end{equation}
%

and aim to approximate the integral of $f(x)$ over the interval $[0,1]$. We construct the initial data $D^{(0)}_{h}$ from a 4-dimensional vector of step-sizes ${\bf h}_{K} = \{1,1/2,1/3,1/4\}$ and use the trapezoidal rule to compute each $I(f, h_{j}), j=1,2,\dots,4$, and set the step-control parameter $\gamma = 0.5$. For the GP, we chose the Mat\'{e}rn class of kernel \cite{Stein1999}:

$$ K_{matern}(r) = \frac{2^{1-\nu}}{\Gamma(\nu)} \bigg( \frac{\sqrt{2 \nu} r}{l} \bigg)^{\nu}  K_{\nu} \bigg( \frac{\sqrt{2 \nu} r}{l} \bigg)$$

where $r=|x-x'|$, $k_{\nu}$ is a modified Bessel function and $\{l,\nu\}$ are the kernel parameters and set $\nu = 1.5$ (for mean square differentiability). We discounted the popular radial basis function (also called squared exponential), which is a limiting case of the Mat\'{e}rn class (as $\nu \to \infty$), as it makes stronger assumptions on smoothness and it has been argued that this may be unrealistic for modeling many physical systems, see \cite{Stein1999,GPbook}. The results from our analysis are presented in Fig. \ref{fig:exp}. We note that a consequence of the GP modeling approach is that the posterior mean of $\hat{I}(f, h)$ is strongly influenced by the choice of prior mean when $h$ lies in the interval between the zero and the most recent update of the step size $h_{K+i}$, i.e. $h\in (0,h_{k+i})$. We therefore assessed the sensitivity of our results to the choice of prior mean across three prior initializations: zero, the average of the $K$ initial points $\sum_{j = 1}^{K} \frac{I(f, h_{K})}{K}$ and $I(f, h_{K})$, i.e. evaluated at the smallest step-size in the initial data set $\mathcal{D}^{(0)}_{h}$ (Fig. \ref{fig:exp} left). In our example, the results illustrate a strong preference toward setting the prior mean as the final observation $\hat{I}(f, h_{4})$. Working under the assumption that the most recent evaluation of I is the most accurate, the result seems reasonable. Fig. \ref{fig:exp} (right) displays the results when applying the $\mathcal{GP}$ approach and demonstrates how the sequential updating of the pair $\{h_{j}, I(f,h_j)\}$, and thus the data $\mathcal{D}_{h}$, can quickly shrink the uncertainty both in the estimation of the integral $F_{\ast}$ at $h=0$. It is possible, therefore, that dynamic updating of the step-size $h_{j}$ to $h_{j+1}$ can lead to faster convergence properties than when fixing the step-size, as in the deterministic set-up.



\begin{algorithm}
    \caption{Richardson extrapolation style algorithm}
  \begin{algorithmic}[1]
    \REQUIRE function $f$, integration interval $H$, step control $\gamma$  where $\gamma \in (0,1$), tolerance $\tau$
    \STATE \textbf{Initialization} evaluate integral with  $h=H,H/2$ 
    \WHILE{not converged}
        \STATE condition $\mathcal{GP}$ on $\mathcal{D}_h$ and calculate confidence ($\sigma_0$) for $h=0$
        \IF{$\sigma_0 < tolerance$} 
        \STATE Accept and exit loop
        \ELSE
        \STATE find $h_i$ with uncertainty $\gamma\sigma_0$
        \STATE evaluate $x_i = I(f,h_i)$
        \STATE append  $\mathcal{D}_h$  with $(h_i,x_i)$
        \STATE repeat
        \ENDIF
    \ENDWHILE
  \end{algorithmic}
  \label{algRE}
\end{algorithm}

\section{Bulirsch-Stoer Algorithm}
\label{sec:headings}
We have reviewed the Bulirsch-Stoer extrapolation algorithm by fixing the domain of integration to be the length the initial step-size $h_{0}$. If the domain is local to a singular point, however, numerical evaluations which rely on a fixed initial step-size can regularly fail to converge. The full implementation of the Bulirsch-Stoer algorithm aims to resolve this issue. In regions where a converged numerical solution appears difficult to identify, the choice of initial step-size, and thus the domain of integration, is repeatedly reduced until a converged numerical solution has been identified. Hence, the original domain of integration is partitioned into segments, in which a converged solution is identified, and the integral is evaluated by aggregating the results from each segment, see Alg. \ref{Alg:BSA} for pseudo code. Our probabilistic re-interpretation of this algorithm sees our GP scheme replacing the traditional extrapolation algorithm, as before, for each segment in the domain, see Alg. \ref{Alg:BSA_GP} for pseudo code. To better suit our numerical example in this section, we henceforth consider the domain of integration as a temporal region and denote a step-size (time-step) by $dt$.   

\begin{algorithm}
    \caption{Bulirsch-Stoer Algorithm}
  \begin{algorithmic}[1]
    \REQUIRE initial conditions $x$, derivative $f'(x)$, tolerance $\tau$, step-size sequence $\{ s_i \}_{i=1}^{n}$
\WHILE{t < H}
    \STATE CONV = FALSE
    \FOR{i <= 8}
        \STATE set step size   $h_i = H/s_i$
        \STATE integrate from $t$ to $t+H$ and store result $R_i$
        \STATE fit polynomial $\{(h_j,R_j)\}_{j=1}^i$ and extrapolate to $h=0$
        \IF{ $|R_i-R_{i-1}|   < \tau $} 
              \STATE CONV = TRUE, accept result and process to next interval 
        \ENDIF
        \ENDFOR
       \IF{ CONV = FALSE} 
            \STATE reduce H and repeat loop
       \ENDIF
           \STATE t += H
    \ENDWHILE
  \end{algorithmic}
  \label{Alg:BSA}
\end{algorithm}

The Bulirsch-Stoer algorithm is descendant of Richardson's deferred approach to the limit and therefore typically\footnote{It is worth noting that the original method as proposed by Bulirsch-Stoer used rational function extrapolation, however straightforward polynomial extrapolation is typically slightly more efficient than rational function extrapolation \cite{NumRce} (for smooth problems).} relies on performing polynomial extrapolation between the numerical evaluations at each abscissa. Briefly, for each segment of integration, the algorithm usually selects the abscissa at the points  $dt/n_{j}$, where $n_{j}$ follows the series originally proposed by Bulirsch-Stoer $n = 2,4,6,...$ $n_j = n_{j-2}$. However, it is more common to see the sequence more recently suggested by Deuflard \cite{Deuflard1,Deuflard2} -- i.e.  $n = 2, 4, 6,... n_j = 2j$  -- which is the scheme we adopt. Typically, the choice of $j=8$ is made \cite{Deuflard1,Deuflard2}. The Bulirsch-Stoer algorithm is an adaptive deterministic numerical scheme, which furnishes solutions to the integral problem across a broad range of scenarios. However, the approach is underpinned by the assumption that the integrand $f(x)$ is well approximated by a polynomial in each segment of the domain of integration which, even for an analytic function $f(x)$, can lead to issues. We now illustrate this. 

\begin{figure}
    \centering
    \includegraphics[width=0.75\textwidth]{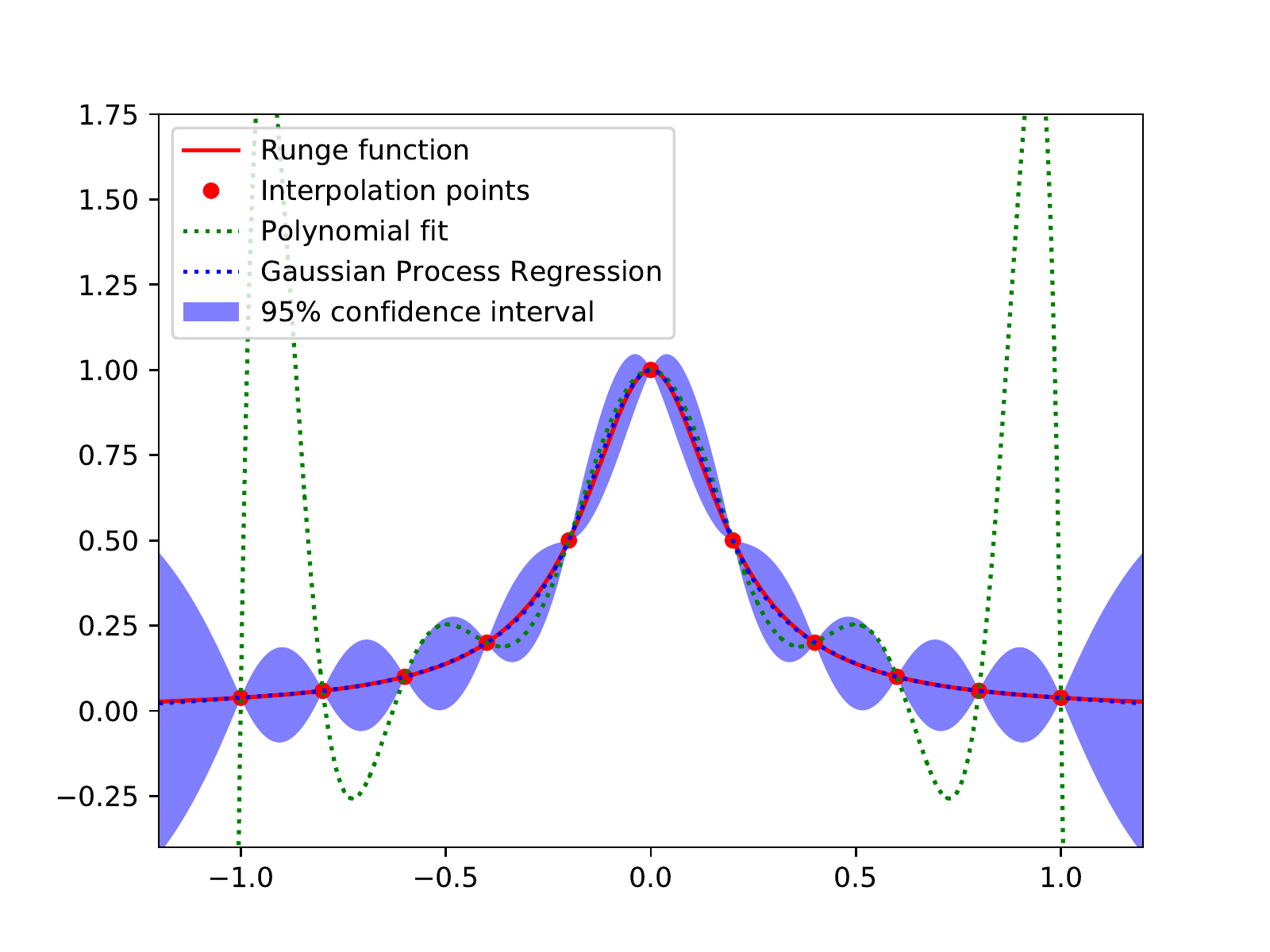}
    \caption{The well-known case of Runge's phenomenon where polynomial interpolation results in oscillation at the edge of the interval. Gaussian Process Regression does not suffer from the same problem and provides a better fit on the same data.}
    \label{fig:runge}
\end{figure}

In Fig. \ref{fig:runge} we provide an example of a well-known situation where polynomial interpolation can perform badly. In this famous example of {\sl Runge's phenomenon}, the GP approach, using a Mat\'{e}rn kernel \cite{Stein1999}, provides excellent extrapolation and interpolation whereas the polynomial interpolation procedure fails, particularly at the tails of the domain. This result might be reversed however, particularly if the true underlying function is or is well approximated by a polynomial. In this case, the GP scheme  will almost certainly fail to match the  goodness of the polynomial extrapolation scheme. The extent to which the methods differ in performance will likely be dependent on the the choice of kernel function: if we had some prior knowledge that the function should `look like' a polynomial, we might choose a kernel that helps the GP posterior mean to exhibit polynomial like behaviour, see for example \cite{Karvonen1998}.

\begin{algorithm}
    \caption{Bulirsch Stoer Algorithm using Gaussian Processes}
  \begin{algorithmic}[1]
    \REQUIRE initial conditions $x$, derivative $f'(x)$, tolerance $\tau$
\WHILE{t < H}
     \STATE evaluate $f$ with step size $H,H/2$ 
    \WHILE{not converged}
        \STATE choose step size $h_i$ as in Alg. \ref{algRE}
        \STATE integrate from $t$ to $t+H$ store result $R_i$
        \STATE append $(h_i,R_i)$ to $\mathcal{D}_h$ 
        \STATE condition $\mathcal{GP}$  on $\mathcal{D}_h$ 
        \STATE evaluate mean value and uncertainty $\sigma_0$ at $h=0$
        \IF{ $\sigma_0  < \tau $} 
              \STATE CONV = TRUE; 
              \STATE break
        \ENDIF
            \STATE repeat loop 
    \ENDWHILE
           \STATE t += H
    \ENDWHILE
  \end{algorithmic}
    \label{Alg:BSA_GP}
\end{algorithm}

In the probabilistic re-interpretation of the Bulirsch-Stoer algorithm, we make the assumption that the numerical evaluation converges to the correct solution as $dt \to 0 $.  This is typically true, but when the domain is partitioned into sufficiently small time-steps rounding error can dominate error due to discretization. The GPR derived scheme suffers from the same issue. Outside of numerical rounding error, the GPR scheme can assess convergence using a combination of two conditions: assessing the uncertainty of the predicted mean in the limit as $dt\to0$, via the fitted GP model, and; by comparing the estimated means between any two intializations. The standard Bulirsch-Stoer algorithm only assesses the later.

The main disadvantage of the probabilistic version of the Bulirsch-Stoer algorithm is computation time, i.e. for a fixed number of abscissa fitting a polynomial using Neville's algorithm is typically faster than aiming to maximize the likelihood of the kernel parameters. Certainly, for a smooth and bounded function $f(x)$, the original BSA will likely outperform the probabilistic version. However, in cases where a segment of the domain is partitioned multiple times, e.g. owing to $|f'(x)|$ rapidly changing over the interval, it is possible that the overall performance of the probabilistic version will be better than the original.

\subsection{Example: Kepler orbit}
To illustrate utility in the probabilistic re-interpretation of the Bulirsch-Stoer algorithm, we consider the case of a mass-less particle in a Keplerian potential \footnote{The choice of an astronomical problem is perhaps fitting as an early reported use of Gaussian processes dates to time series analysis by the astronomer T.N. Thiele in 1880. \cite{L1981}}. The problem involves finding a solution to the second order ODE, 

$$ \ddot{\bf r} = \frac{GM}{|{\bf r}|^3 } {\bf r} $$

where ${\bf r}$ is the position vector, $G$ is Newton's constant and $M$ is total mass (it is convention to use units with $G=M=1$). We consider the case of a particle initially at ${\bf r} = (1,0)$ on an orbit with eccentricity of 0.99. This results in a close passage to the singularity in the force at ${\bf r} = {\bf 0}$.

We compare two different versions of Bulirsch-Stoer: the usual polynomial extrapolation and; the Gaussian process regression approach. Integration over each sub-interval is performed using a second-order leapfrog integrator. For the polynomial scheme, if integration fails to converge after eight sub-divisions of the time-step $dt/n{j}$, the region is iteratively halved until convergence is achieved. We note that more complex schemes can be used to reduce the time-step  (e.g. see \cite{NumRce}), which might outperform the Bulirsch-Stoer scheme. For the GP approach, we do not place a limit on the number of times an interval can be sub-divided owing to the ability of the GP scheme to closely approximate a wider class of analytic functions than those that are well approximated by a finite order polynomial. 

\begin{figure}
    \centering
    \includegraphics[width=0.75\textwidth]{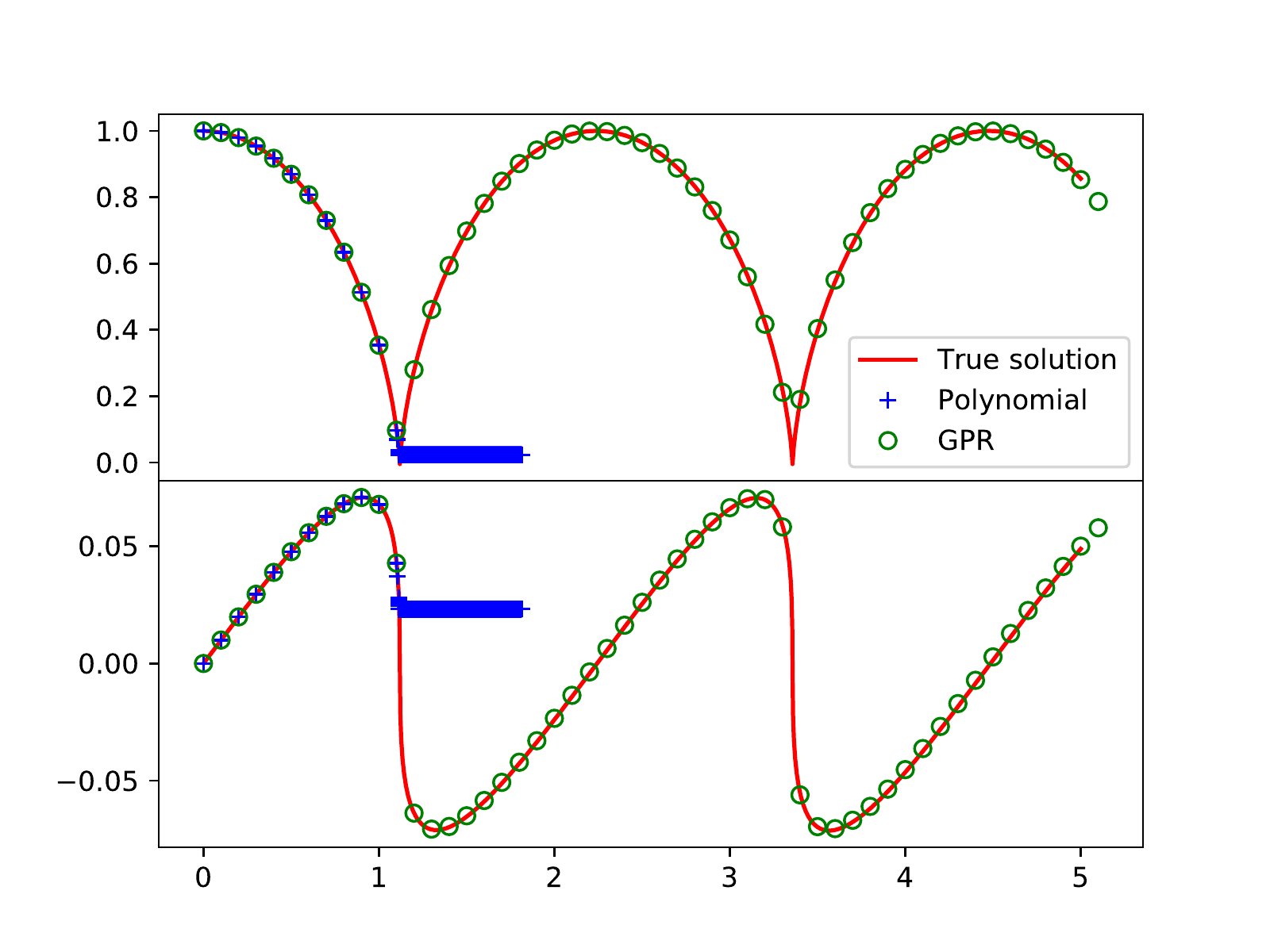}
    \caption{Comparison of Bulirsch-Stoer schemes with polynomial extrapolation and Gaussian process regression. Displayed are the x (horizontal) and y (vertical) co-ordinates of a particle, as a function of time, in a Kepler orbit with eccentricity of 0.99. The Bulirsch-Stoer scheme using polynomial extrapolation (blue circles) fails during the pericenter passage (i.e. closest approach). In contrast, Gaussian process regression (green circles) closely folows the true particle trajectory across the spatio-temporal domain.}
    \label{fig:kepler}
\end{figure}

Fig. \ref{fig:kepler} displays a comparison of the two schemes relative to the true solution. Notably, as the particle first passes through pericenter passage, i.e. the closest approach to the origin, which is a singularity, the polynomial based scheme fails and no longer furnishes an estimate of the particle's position. However, the non-parametric GP scheme closely approximates the particle's trajectory as it first passes through pericenter and, moreover,continues to furnish a reasonable approximation of the particle path throughout the time interval considered.

\section{Discussion}
In this paper we have presented an adaptive-learning integration scheme, which embeds Gaussian process regression (GPR) into the Bulirsch-Stoer algorithm (BSA), for use in numerical integration. This probabilistic approach extends the current deterministic implementation of the BSA in two ways. Firstly, the approach avoids the parametric assumption that a numerical evaluation of the integral has an error term that is closely approximated by a finite order polynomial. This assumption can lead to the deterministic approach failing near a singular point in the integrand. Secondly, for given step-size, uncertainty in the numerical evaluation of the integral via GPR is used to guide how the step-size should be adapted. The combination of these additional features leads to a more robust numerical integration scheme beyond the current standard. The probabilistic scheme is not intended one-size-fits-all replacement for existing numerical methods, however, but rather to be considered as an alternative when standard approaches fail.

\subsubsection*{Acknowledgments}

We would like to thank Prof. Chris Oates for helpful comments in connection with the project and Dr. Paul Kirk for interesting scientific discussions and helpful comments  on the manuscripts. PB acknowledge support from the Leverhulme Trust
(Research Project Grant, RPG-2015-408). CNF acknowledges support from the Medical  Research  Council  (MC  UU  00002/7).


\end{document}